\documentclass{article}

\usepackage{microtype}
\usepackage{graphicx}
\usepackage{subfigure}
\usepackage{booktabs} 
\usepackage{amsmath}
\usepackage{epstopdf}

\usepackage{hyperref}

\usepackage[accepted]{icml2018}

\icmltitlerunning{Multiview Representation Learning for a Union of Subspaces}

\usepackage{amssymb}

\newcommand{\U}{\textrm{U}}
\newcommand{\V}{\textrm{V}}
\newcommand{\I}{\textrm{I}}
\newcommand{\C}{\textrm{C}}
\newcommand{\E}[1]{\underset{#1}{\mathbb{E}}} 
\newcommand{\tr}{\textrm{tr }} 
\newcommand{\Rm}[2]{\mathbb{R}^{#1 \times #2}} 
\newcommand{\R}[1]{\mathbb{R}^{#1}} 

\begin{document}

\twocolumn[
\icmltitle{Multiview Representation Learning for a Union of Subspaces}



\begin{icmlauthorlist}
\icmlauthor{Nils Holzenberger}{jhu}
\icmlauthor{Raman Arora}{jhu}
\end{icmlauthorlist}

\icmlaffiliation{jhu}{Department of Computer Science, Johns Hopkins University, Baltimore, MD, USA}

\icmlcorrespondingauthor{Nils Holzenberger}{nholzen1@jhu.edu}

\icmlkeywords{Machine Learning, ICML}

\vskip 0.3in
]



\printAffiliationsAndNotice{}  

\begin{abstract}

Canonical correlation analysis (CCA) is a popular technique for 
learning representations that are maximally correlated across multiple views in data. In this paper, we extend the CCA based framework for learning a multiview mixture model. We show that the proposed model 
and a set of simple heuristics yield improvements over standard CCA, as measured in terms of performance on downstream tasks. Our experimental results 
show that our correlation-based objective meaningfully generalizes the CCA objective to a mixture of CCA models. 

\end{abstract}

\section{Introduction}
\label{sec:introduction}

Multiview, correlation-based representation learning has been shown to be useful on a variety of tasks \citep{hardoon2004canonical,wang2015unsupervised,arora2013multi,arora2014multi,benton2016learning,vasquez2017multi,holzenberger2019learning}. Its main workhorse, Canonical Correlation Analysis (CCA) \citep{hotelling1936relations}, enjoys several non-linear extensions with theoretical guarantees (kernel CCA and deep CCA) \citep{wang2015deep, lai2000kernel}, and can be extended to more than 2 views \citep{horst1961generalized,rastogi2015multiview,benton2017deep}. 

A related approach that yields an arguably richer representation is to learn a mixture model --- instead of learning a single, high-dimensional, complex subspace, we can seek to find a union of subspaces. Mixture models, especially Gaussian Mixture Models, have a wide range of applications, from topic modeling \cite{blei2003latent} to speech recognition \cite{gales2008application}. In many cases, data can naturally be described as a union of distributions. For instance, phonemes in speech are a superposition of low-dimensional processes, either explained as templates of time-frequency vectors, or as typical articulatory motions \cite{sugamura1983isolated}. In the domain of information extraction, documents fall into categories such as newswire, blog posts, agency reports or tweets. In machine translation, domain match or mismatch plays an important part in the performance of a translation system \cite{koehn2017six}. Identifying the underlying components of the mixture, either explicitly or implicitly, can help unsupervised modeling of the data, useful for a host of  downstream tasks.

Learning a mixture of subspaces rather than a single subspace also makes personalization simpler. For an unseen example, being able to assign it quickly to a subpopulation can make classification tasks require less data to achieve good performance. For example, in the context of speaker adaptation, being able to assign a speaker to a specific group of speakers improves speech recognition performance \cite{kuhn2000rapid}.

In this paper, we propose to extend the framework of CCA to learn a union of subspaces. Specifically, we assume that each data point belongs to one of a finite number of sources, and we learn a pair of linear CCA transformations for each source. CCA with a single transformation can detect canonical directions that span the entire dataset. A mixture of CCA is also able to detect the main canonical directions, because they can be found in each subpopulation. In addition, a mixture of CCA can pick up correlations which are present at the subpopulation level but cancel each other out at the population level. This argument would suggest we can provide a much finer grained representation, possibly without increasing the dimensionality of the representation.

While the premise might seem straightforward, combining CCA and mixture models poses a number of challenges: what objective is to be maximized? How to simultaneously learn the cluster assignments and the transformations? Are there any guarantees regarding convergence, and recovery of cluster memberships? Once the parameters of the mixture model are learned, how to assign a new point $(x,y)$ to a cluster? How to assign a single $x$ without corresponding $y$ to a cluster? This paper is meant primarily as a proof of concept, and leaves most of the above questions open. We propose a new objective function, as well as a heuristic way of maximizing it, and test its performance in two distinct settings.

\section{Related work}
\label{sec:related_work}

We use Canonical Correlation Analysis (CCA) \cite{hotelling1936relations} to learn representations for a primary view. We assume that at training time, we are given two views of the same data point. For instance, for a given speech utterance, the audio recording and the articulatory measurements. These two views are represented by random variables $X$ and $Y$ ($d_X$- and $d_Y$-dimensional respectively). Linear CCA seeks two linear transformations $\U \in \Rm{d_X}{k}$ and $\V \in \Rm{d_Y}{k}$ such that the components of $\U^TX$ and $\V^TY$ are maximally correlated. Formally, we want to maximize $\E{X,Y} [ \tr(\U^TXY^T\V) ]$ subject to the constraints that ${\E{X} [ \U^TXX^T\U ] = \E{Y} [ \V^TYY^T\V ] = \I_k}$.

Given a dataset $\{x_i,y_i\}_{i=1}^N$, we define $\C^{XY}$ the empirical cross-covariance matrix between $X$ and $Y$, and $\C^{XX}$ and $\C^{YY}$ the empirical auto-covariance matrices of $X$ and $Y$, respectively. $\U$ and $\V$ are given by the $k$ left and right singular vectors of $(\C^{XX})^{-1/2}\C^{XY}(\C^{YY})^{-1/2}$ with the largest singular values, multiplied by $(C^{XX})^{-1/2}$ and $(C^{YY})^{-1/2}$.

At test time, we assume that only the primary view is available. Ideally, we would want to reconstruct the second view with the primary view \cite{ngiam2011multimodal}. However, in general, that is a difficult task --- take for example generating speech from text. Instead, it is easier to predict the dependent variate which has the largest multiple correlation. In other words, when no single regression provides a fully adequate solution, CCA is a better objective than predicting one view with the other.

Deep CCA (DCCA) \cite{andrew2013deep} is a natural extension of linear CCA, where one seeks to maximally correlate $\U^Tf(X)$ and $\V^Tg(Y)$. $f$ and $g$ are non-linear feature extractors, which can be learned via gradient descent on the CCA objective. It is also natural to extend CCA to multiple views \cite{horst1961generalized}. 

CCA can also be reformulated as a probabilistic model. \citet{bach2005probabilistic} show that the solution to the linear CCA objective function is, up to arbitrary rotation and scaling, contained in the maximum-likelihood solution for the parameters of a Gaussian. \citet{podosinnikova2016beyond} extend this probabilistic formulation of CCA to fit any type of statistical distribution, including when one or both views are discrete.

\citet{klami2007local} place the probabilistic model of \citet{bach2005probabilistic} in a Bayesian setting, with a Dirichlet process, which naturally leads to a mixture of Gaussians. They then maximize log-likelihood under this GMM model using methods from variational inference. As a final step, they extract CCA projections from each Gaussian. This amounts to performing GMM-based clustering first, and then performing CCA on each of the learned clusters. In contrast, we define a correlation-based objective for multiple transformations, and propose to maximize it directly. It remains to be examined whether both objective functions are equivalent, or even whether stationary points of both objective functions are equivalent. In fact, the model of \citet{klami2007local} might be suboptimal in terms of our objective. Other works \citep{wang2007variational,viinikanoja2010variational,hosino2010high} have used very similar approaches.

Most related to this work is that of \citet{fern2005correlation}, where the authors use of a mixture of CCA, and provide an optimization heuristic similar to ours. However, their work differs on a number of points. First, \citet{fern2005correlation} do not provide an objective function to optimize. Second, they do not test the CCA projections in a downstream task, and are thus arguably not concerned with questions pertaining to representation learning. Third, \citet{fern2005correlation} do not provide any way of assigning a point with a single view to a cluster (i.e., they specify how to assign $(x,y)$ to a cluster, but not how to assign $x$ to a cluster in the absence of $y$). The main focus of our work is the objective function, as a possible extension of CCA and representation learning method.

\section{Mixture model for Canonical Correlation Analysis}

We propose to learn a union of subspaces that best characterizes a set of points with two views, using canonical correlation analysis (CCA). It involves assigning cluster memberships to data points, and we limit ourselves to linear versions of CCA. Throughout this paper, we will refer to this method as mixture of CCA (MCCA).

\subsection{Objective formulation}
\label{subsec:objective_formulation}

We assume that our data consists of two sets of points.
Let $X \in \Rm{d_X}{N}$, $Y \in \Rm{d_Y}{N}$ be the matrices containing the paired views, with $x_i \in \R{d_X}$ (resp. $y_i \in \R{d_Y}$) being the $i$-th column of $X$ (resp. $Y$). For example, in our application to speech recognition, $X$ represents a set of MFCC frames and $Y$ the corresponding articulatory measurements. We consider a mixture of $R$ subspaces and consider the following CCA formulation. We define the MCCA objective as:

$$O(U,V,\alpha) = \sum \limits_{r=1}^R \tr(U_r^T C^{XY}_r V_r)$$

subject to the constraints that

$${\forall r \in \{1..R\}, \thickspace U_r^T C^{XX}_r U_r = V_r^T C^{YY}_r V_r = \I_k}$$

where ${U=\{U_r \in \Rm{d_X}{k} \mid r=1,\ldots,R\}}$, and ${V = \{V_r \in \Rm{d_Y}{k} \mid r=1,\ldots,R\}}$ are the CCA transformation matrices associated with each mixture~component. 

The assignment of data points to mixture components is given by the following scalar weights:

$${\alpha = \{\alpha_{i,r} \in \mathbb{R}_{\geq 0} \mid i \in [N], r\in [R], \forall i, \sum \limits_{r=1}^R \alpha_{i,r} = 1\}},$$

where $[p]$ denotes the set $\{1,\ldots,p\}$.

The weighted empirical covariance and cross-covariance matrices for each mixture component are given as

$$C^{XY}_r = \frac{1}{\sum \limits_{i=1}^N \alpha_{i,r}} \sum \limits_{i=1}^N \alpha_{i,r} (x_i - \mu^{X}_r) (y_i - \mu^{Y}_r)^T,$$

$$C^{XX}_r = \frac{1}{\sum \limits_{i=1}^N \alpha_{i,r}} \sum \limits_{i=1}^N \alpha_{i,r} (x_i - \mu^{X}_r) (x_i - \mu^{X}_r)^T,$$
and

$$C^{YY}_r = \frac{1}{\sum \limits_{i=1}^N \alpha_{i,r}} \sum \limits_{i=1}^N \alpha_{i,r} (y_i - \mu^{Y}_r) (y_i - \mu^{Y}_r)^T,$$

and the weighted means of the mixture components are given as

$$\mu^{X}_r = \frac{1}{\sum \limits_{i=1}^N \alpha_{i,r}} \sum \limits_{i=1}^N \alpha_{i,r} x_i,~~~ \mu^{Y}_r = \frac{1}{\sum \limits_{i=1}^N \alpha_{i,r}} \sum \limits_{i=1}^N \alpha_{i,r} y_i.$$ 

At training time, the goal is to find

$$U^*, V^*, \alpha^* = \underset{U,V, \alpha}{\textrm{argmax }} O(U,V, \alpha)~~~\textrm{(subject to constraints)},$$
i.e., to find for each mixture component, both CCA transformation matrices and the assignments for each data point $(x_i,y_i)$.

At test time, given $x$ and/or $y$, one needs to estimate the $\{\alpha_r\}_r$ before projecting $x$ (respectively, $y$) to $\sum \limits_{r=1}^{R} \alpha_r U_r^T x$ (respectively, to $\sum \limits_{r=1}^{R} \alpha_r V_r^T y$).

To avoid spurious correlations, we regularize the covariance matrices by adding scaled identity matrices, i.e., replacing $C^{XX}$ (respectively, $C^{YY}$) with $C^{XX} + w_X I$ (respectively, $C^{YY} + w_Y I$). 

\subsection{Learning and inference}
\label{subsec:learning_inference}

The optimization problem in Section~\ref{subsec:objective_formulation} is nonconvex, and jointly minimizing over all parameters seems daunting from a computational perspective due to intractability. Therefore, we consider the following optimization approach that first estimates $\alpha$, then estimates $U$ and $V$ given $\alpha$, as summarized in Algorithm \ref{alg:mcca_alternate_maximization}. When $\alpha$ is fixed, it is well known that we can globally maximize the $R$ CCA objectives over the choice of $U$ and $V$, as described in Section~\ref{sec:related_work}. Similarly, fixing the CCA subspaces, maximizing over $\alpha$ is essentially learning a mixture model over the shared representation.

\begin{algorithm}[tb]
   \caption{Heuristic optimization for MCCA}
   \label{alg:mcca_alternate_maximization}
\begin{algorithmic}
   \STATE {\bfseries Input:} MCCA hyperparameters $\{R,k,w_X,w_Y\}$ and dataset $\{X \in \Rm{d_X}{N}, Y \in \Rm{d_Y}{N}\}$
   \STATE {\bfseries Output:} Transformation matrices $U$ and $V$, and cluster assignments $\alpha$ 
   \STATE initialize $\alpha$ using a clustering algorithm over the native space
   \STATE find $U,V = \underset{\tilde{U}, \tilde{V}}{\textrm{argmax }} O(\tilde{U}, \tilde{V}, \alpha)$
   \STATE return $U, V, \alpha$
\end{algorithmic}
\end{algorithm}

At training time, we first cluster the data points using a CCA projection and k-means clustering, yielding hard assignments for $\alpha$, i.e. ${\alpha \in \{0,1\}^{NR}}$. We then learn $U$ and $V$.

At test time, we need a way of inferring the $\{\alpha_{i,r}\}_{r}$ for a given $(x_i,y_i)$ where $x_i$ or $y_i$ may be missing. In our experimental setting, only $x$ is available at test time. From the properties of the CCA projections, we know that $\mathrm{var}(U^TX) = I_k$. Heuristically, we expect $U^T (x-\mu_X)$ to be drawn from a unit variance Gaussian distribution. Thus, we assign $x$ to the cluster 
$$\hat{r} = \underset{r \in \{1,\ldots,R\}}{\textrm{argmin}} ||U_r^T(x-\mu^X_r)||_2^2 - \log \pi_r,$$ 
where $\pi_r$ is the fraction of points belonging to cluster $r$. In some of our speech recognition experiments, $\alpha$ will be given at training time.

\section{Experiments}

In this section, we illustrate the possible uses of MCCA, and test its performance against standard, vanilla linear CCA described in Section~\ref{sec:related_work}, and referred to as VCCA. All non-linear extensions of CCA -- deep CCA and kernel CCA -- have significantly more representational power than linear CCA. Most likely, deep CCA would outperform all other methods. Thus, for the comparison to non-linear extensions of CCA to be fair, one would have to extend MCCA to non-linear methods. This would pose many more challenges related to optimization. Instead, our intention is to carefully illustrate the linear version, leaving non-linear extensions to future work.

\subsection{Phoneme classification}
\label{subsec:phoneme_classification}

The University of Wisconsin X-ray microbeam database (XRMB) \cite{westbury1990x} is a set of sound recordings and articulatory measurements, acquired during production of English read speech. The acoustic recordings are processed into sequences of 13 dimensional MFCC frames. The articulatory measurements are the horizontal and vertical displacements of eight pellets affixed to critical articulators (tongue, lips, jaw) at a given point in time, resulting in a 16 dimensional observation vector. Both views are temporally aligned, and thus we have pairs of MFCC frames and articulatory features. This correspondence is the only supervisory signal when learning representations; while the manual annotations of the spoken text are available, we do not use it for feature learning to avoid any task dependence in the learned representations. The location of the pellets in the original dataset go missing for various reasons; we use the completed version of the XRMB dataset due to \citet{wang2014reconstruction}. The MFCC frames are augmented with deltas and double deltas, and are mean-centered and variance-normalized per speaker. The articulatory features are also mean-centered and variance-normalized per speaker.

In this section, we illustrate the usefulness of learning a union of subspaces, by partitioning the 39 phonemes of the XRMB dataset into 4 groups, as detailed in Table~\ref{tab:xrmb_groups}. For some experiments, we assume access at training time to an oracle providing us with the correct group for each training instance; in those cases, we are not concerned with inferring $\alpha$. At test time, we also experiment with the presence and absence of said oracle. In the absence of the oracle, we use the heuristic described in Section~\ref{subsec:learning_inference} to infer the membership of each point.

\begin{table}[t]
\caption{4 groups used to partition the English phonemes.}
\label{tab:xrmb_groups}
\vskip 0.15in
\begin{center}
\begin{small}
\begin{sc}
\begin{tabular}{ p{1.5cm} p{5.5cm} }
\toprule
vowels & AA, AH, IH, AO, ER, EH, IY, UW, AE, OW, UH, AY, EY, AW, OY, HH, Y \\
alveolars & CH, D, DH, JH, L, N, R, S, SH, T, TH, Z, ZH \\
labials & B, F, M, P, V, W \\
velars & G, K, NG \\
\bottomrule
\end{tabular}
\end{sc}
\end{small}
\end{center}
\vskip -0.1in
\end{table}

Using the notation of Section~\ref{subsec:objective_formulation}, $X$ consists of 7 stacked MFCC frames with deltas and double deltas, centered around the frame of interest, and is thus 273 dimensional vectors. $Y$ consists of 7 stacked articulatory feature vectors, corresponding to the 7 MFCC frames, and is thus 112 dimensional. The training set has 1.4M data points, and each of the 6 cross-validation folds mentioned below has between 79k and 86k data points.

To evaluate the quality of the learned representations, we follow \citet{wang2015unsupervised} and leave out 12 speakers. The other 35 speakers are used to learn representations. The 12 left-out speakers are used to measure how well the learned representations for audio can be used to perform phoneme classification on unseen data. To estimate generalization error, the 12 speakers are partitioned into 6 sets of 2 speakers each, and used in a 6-fold cross-validation, each fold composed of 4 training speakers, 2 dev speakers, and 2 test speakers. We use the 4 train and 2 dev speakers to find the best parameters for a k-nearest neighbor classifier \cite{malkov2018efficient}, then measure the score on the dev set. We use the average dev score over the 6 folds to compare hyperparameters of a given method. This procedure allows us to pick the best hyperparameters for a given CCA method (VCCA or MCCA). Methods are then compared on the 6 test sets, with hyperparameters $\Theta^*$ and $\Phi^*$. We report average and standard deviation of the performance over the 6 dev and test sets. Knn classifier hyperparameters $\Phi$ contain the type of distance used (L2 or cosine), the number of neighbors (8, 16, 32, 64, 128 or 256), and whether to append the original MFCC features to the CCA features.

 We sweep $k$ over $\{10, 30, 50, 70, 90, 110\}$, and $w_X$ and $w_Y$ independently over $\{1, 0.1, 0.001\}$. For VCCA, we use the projection of the MFCC features, and experiment with appending the original MFCC features to perform knn classification. For MCCA, we also experiment with appending the original MFCC features. In addition, when projecting point $x$, we have the possibility of mapping $x$ to $\sum \limits_{r=1}^R \alpha_{r}U_r^T x \in \R{k}$ or to the concatenation of $U_1^Tx, U_2^Tx, ... , U_R^T x$, which is an $Rk$ dimensional vector. Note that the former requires inferring $\alpha$ at test time, while the latter doesn't. We experiment with both settings, reporting the former as ``projection'' and the latter as ``concatenation''.

Table~\ref{tab:xrmb_results_noracle} reports our experimental results. We experiment with the oracle of Table~\ref{tab:xrmb_groups} being present or absent at training time (denoted by ``oracle'' and ``no oracle'' respectively). When the oracle is absent at training time, we sweep $R$ over $\{2, 4, 8, 16\}$. In this setting, both instances of MCCA yield a significant improvement over VCCA. The best dev score for VCCA is achieved with $k=70, w_X=0.001, w_Y=1$, while the best score for MCCA with oracle is achieved with $k=50$ and the same values of $w_X$ and $w_Y$; the best score without oracle is achieved with $k=30$, $w_X=w_Y=0.001$ and $R=8$. There is no reduction in the dimensionality of the representation because the best dev score is achieved by concatenating all $R$ representations.

\begin{table}[t]
\caption{Phoneme classification accuracy on XRMB dataset without oracle at test time. Oracle (resp. no oracle) indicates access (resp. no access) to the oracle at test time.}
\label{tab:xrmb_results_noracle}
\vskip 0.15in
\begin{center}
\begin{small}
\begin{sc}
\begin{tabular}{ p{4cm} p{1.5cm} p{1.5cm} }
\toprule
features & dev (\%) & test (\%) \\
\midrule
MFCC & $60.6 \pm 2.8$  & $60.6 \pm 2.5$ \\
VCCA & $65.3 \pm 2.8$ & $65.3 \pm 2.9$ \\
MCCA (oracle, projection) & $64.1 \pm 2.7$ & $64.2 \pm 2.6$ \\
MCCA (oracle, concatenation) & $68.0 \pm 2.7$ & $68.0 \pm 2.8$ \\
MCCA (no oracle, projection) & $64.7 \pm 2.8$ & $64.8 \pm 2.7$ \\
MCCA (no oracle, concatenation) & $69.2 \pm 2.8$ & $69.3 \pm 2.8$ \\
\bottomrule
\end{tabular}
\end{sc}
\end{small}
\end{center}
\vskip -0.1in
\end{table}

The results in Table~\ref{tab:xrmb_results_noracle} show that MCCA yields a significant improvement over VCCA, with and without oracle at training time. In practice, it is likely that there is no oracle at training time, either because of a lack of labels or because there is no clear partitioning of the data. The bottom two lines in Table~\ref{tab:xrmb_results_noracle} show that the absence of the oracle does not necessarily imply a loss in performance. In fact, learning each point's assignment to a mixture component yields better results than following the heuristic from Table~\ref{tab:xrmb_groups}.

If the oracle had been available at test time in addition to training time, the instance of MCCA projection reported in Table~\ref{tab:xrmb_results_noracle} would reach $77.5 \pm 2.2 \%$ on the dev set and $77.5 \pm 2.1 \%$ accuracy on the test set. This shows that while our test-time heuristic is able to outperform standard CCA, there is still room for improvement. It also shows that, although our MCCA projection results are below our MCCA concatenation results, given correct point assignments, projection is enough to guarantee a useful representation.

\begin{figure}[ht]
\vskip 0.2in
\begin{center}
\centerline{\includegraphics[width=\columnwidth]{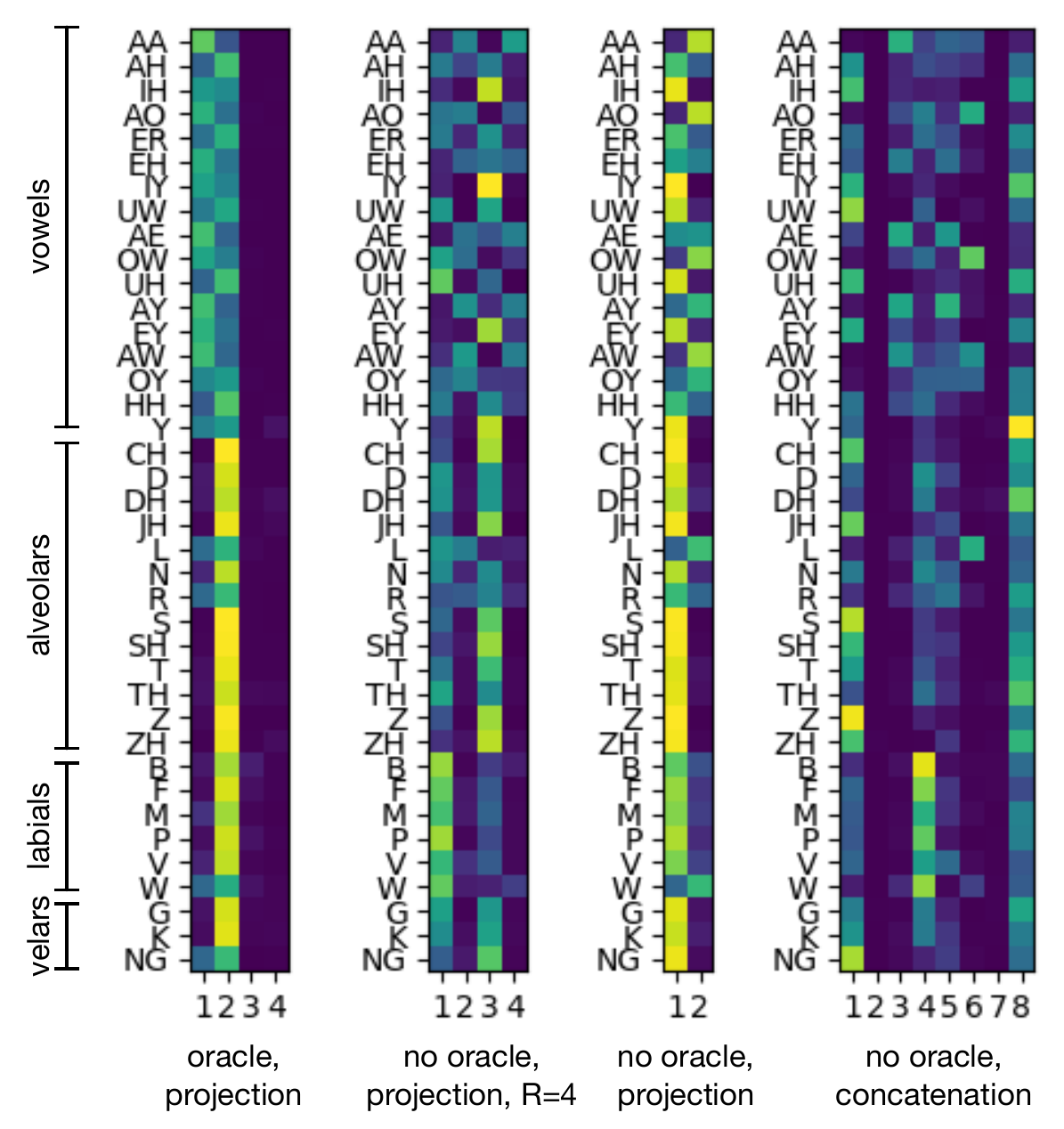}}
\caption{Perplexity matrices for various MCCA models on the test set. Rows are labeled with phonemes and columns with mixture components. Rows are normalized to sum to 1. Each model was chosen as the best in its category, as described by the label. Columns 1, 3 and 4 correspond to the models reported in Table~\ref{tab:xrmb_results_noracle}.}
\label{fig:perplexity_matrices}
\end{center}
\vskip -0.2in
\end{figure}

In Figure~\ref{fig:perplexity_matrices}, we show perplexity matrices for some of our MCCA models on the test set. On that plot, phonemes are sorted according to their membership in the 4 groups of Table~\ref{tab:xrmb_groups}. The first column has access to the oracle at training time, and achieves the best score when projecting points with a single CCA transformation. In that case, ideally, we would want the assignments to be 4 disjoint bands following the 4 clusters. However, each phoneme is mostly assigned to groups 1 and 2, which are also the groups with the largest mixing weights. This mis-assignment can be somewhat alleviated by removing the $\log \pi_r$ term (see Section~\ref{subsec:learning_inference}), but worsens the dev score. Columns 2 shows that, given 4 clusters but no access to the oracle, the clustering heuristic used comes up with sharp clusters for consonants. Roughly, cluster 1 contains the labials, cluster 3 contains the alveolars, and the velars are spread out among clusters 1 and 3. Consonants are mostly spread out between clusters. If we do not constrain the number of clusters to be 4, the best performing model without oracle at training time and using projection has 2 clusters, shown in the third column. In that case, we see mostly sharp assignments, i.e. almost all phonemes are assigned to a single cluster, despite no access to any phonemic information. With some exceptions, consonants belong to cluster 1, and vowels fall into either cluster. The fourth column shows cluster assignments for the best performing MCCA model; note that this model concatenates representations, and thus cluster assignments do not matter at test time. Clusters 2 and 7 seem to be unused, and cluster 4 loosely corresponds to labials. Alveolars and velars are spread out among clusters 1 and 8.

\subsection{Twitter data}

To provide an illustration of MCCA in a different domain, we use the corpus of Twitter data of \citet{benton2016learning}. In that paper, various methods are used to build representations of Twitter users, based on their tweets and friend networks. Following the nomenclature of \citet{benton2016learning}, we use the ego view, i.e. a PCA-based representation of the user's tweets, and the friends view, i.e. a PCA-based representation of the user's friend network or graph. We use the ego view as the primary (available at training and test time) and the friends view as the secondary view (available at training time only). Using the notation of \ref{subsec:objective_formulation}, $X$ (resp. $Y$) is the primary (resp. secondary) view. Both are 1000 dimensional. To estimate how much the use of CCA is improving the performance on a given task, we also report results using the primary view at training time, under ``raw features''.

\subsubsection{User engagement prediction}
\label{subsec:user_engagement_prediction}

To evaluate the learned representations, we perform a user engagement prediction task, using hashtag as a proxy (following \citet{benton2016learning}). We have 2 sets (dev and test) of 200 unseen hashtags each, and for each hashtag, a set of 16k unseen users who have used them. The task is to predict, for each hashtag, which users are likely to use it, based on the first 10 users who have used it. We follow the setup of \citet{benton2016learning}, but report slightly different metrics. Note that because of missing views, we have had to restrict the number of users, including in the test set, and thus the results are not comparable. The train, dev and test set contain respectively 79900, 8220 and 8071 users. For a given method, we use the performance on the dev set to select the best hyperparameters, then compare methods on the test set. We decide whether or not to concatenate the original representation to the CCA projection based on the dev set performance. When concatenating, we scale each representation by its average norm taken over the dataset, in order to balance out the weight of each representation in the cosine similarity.

More specifically: we first project all the users $u_1, u_2, ..., u_N$ of the dev (resp. test) set using the learned CCA transformations. For each hashtag $i$ in the dev (resp. test) set, we have a list of users who have used them. We pick the first 10, and compute the average of their embeddings; we take this as the representation $h_i$ of the hashtag. Hashtag representations $\{h_i\}_i$ and user representation $\{u_j\}_j$ are separately mean-centered. Then, for each hashtag $h_i$ and each user $u_j$, we compute the scaled cosine similarity $\frac{1}{2}(1+\frac{h_i\cdot u_j}{||h_i||||u_j||})$, yielding a confidence score between $0$ and $1$. We use that score as a measure of how likely the user $j$ would use the hashtag $i$. We can compare these scores with the ground truth of which user actually used each hashtag.

We report Recall@1000, mean reciprocal rank, and the area under the ROC curve (ROC-AUC) \cite{fawcett2004roc} of a multiclass classifier mapping users to hashtags. Recall@1000 and ROC-AUC are bounded between $0$ and $100$, and mean reciprocal rank between $0$ and $1$. For all of these metrics, higher is better. For each hashtag, we rank the users based on their confidence score, in decreasing order. Recall@1000 tells us how much the top 1000 users overlap with the ground truth, and so how many relevant answers are present in our top 1000 answers. Mean reciprocal rank tells us how close the first relevant result is to the top of our result list. If we were to build the simplest classifier for hashtag $j$, it would rule that user $i$ is going to use hashtag~$j$ if $\frac{1}{2}(1+\frac{h_i\cdot u_j}{||h_i||||u_j||})>\theta_j$ where $\theta_j$ has to be set for each hashtag. The ROC-AUC tells us how well the confidence scores of correct users are separated from those of incorrect users. A ROC-AUC score of $100$ means we could set the $\{\theta_j\}_j$ randomly and still classify each user and hashtag correctly, and $50$ means we are doing as well as a random classifier. The higher the ROC-AUC score, the better the performance of the classifier will be, regardless of the values of the $\{\theta_j\}_j$.

In contrast to the experiments in Section~\ref{subsec:phoneme_classification}, there is no obvious partitioning of Twitter users on this dataset, and thus we have no oracle at all in this case. We sweep $k$ over $\{200, 400, 600, 800, 1000\}$, and $w_X$ and $w_Y$ independently over $\{1, 0.1, 0.01\}$. In the case of MCCA, we sweep the number of clusters $R$ over $\{2,4,8,16\}$ and follow Section~\ref{subsec:learning_inference} for learning and inference.

We summarize the results of the user engagement prediction task in Table~\ref{tab:twitter_hashtag}. On all three metrics considered, MCCA outperforms VCCA. For Recall@1000 and ROC-AUC, the hyperparameters achieving the best scores are identical. For VCCA, ${k=600,w_X=0.1,w_Y=1}$; for MCCA projection, ${R=2,k=1000,w_X=0.1,w_Y=0.001}$; for MCCA concatenation, ${R=8,k=400,w_X=0.001,w_Y=1}$. For all three, the best scores are achieved with the CCA representation alone (without appending the raw features). In the case of mean reciprocal rank, the best scores are achieved by concatenating the CCA representation and the raw features. For VCCA, ${k=400,w_X=0.001,w_Y=1}$; for MCCA projection, ${R=8,k=800,w_X=0.001,w_Y=1}$; for MCCA concatenation, ${R=2,k=1000,w_X=0.001,w_Y=1}$. The large difference in best performing hyperparameters shows how important task-based hyperparameter selection is. In that respect, MCCA has 2 more hyperparameters that can be tuned based on the task: the number of mixture components $R$, and whether to project $x$ to ${\sum \limits_{r=1}^R \alpha_r U_r^T x}$ (``projection'') or to the concatenation of ${U_1^Tx, U_2^Tx, ... , U_R^Tx}$ (``concatenation'').

Except for mean reciprocal rank, MCCA projection performs consistently worse than VCCA. This could possibly be because the clusters picked by our heuristic have no intrinsic relationship with the hashtags. It is interesting to note that MCCA projection drastically improves the mean reciprocal rank to perfect or almost perfect score, meaning that the top result is a correct answer, almost always. Higher ROC-AUC scores for MCCA concatenation show that this method is better able to separate, for each hashtag, relevant from irrelevant users. This is consistent with higher Recall@1000 scores, which indicate more relevant users within the top 1000 results.

\begin{table}[t]
\caption{User engagement prediction results on Twitter dataset. REC: Recall@1000, ROC-AUC: area under the ROC curve, MRR: mean reciprocal rank.}
\label{tab:twitter_hashtag}
\vskip 0.15in
\begin{center}
\begin{small}
\begin{sc}
\begin{tabular}{ p{2cm} p{0.5cm} p{0.5cm} p{0.5cm} p{0.5cm} p{0.5cm} p{0.5cm} }
\toprule
& \multicolumn{2}{c}{REC} & \multicolumn{2}{c}{ROC-AUC} & \multicolumn{2}{c}{MRR} \\ 
\cmidrule(r){2-3}\cmidrule(r){4-5}\cmidrule(r){6-7}
features & dev & test & dev & test & dev & test \\
\midrule
raw features & 53.0 & 44.0 & 78.4 & 72.0 & .858 & .876 \\
VCCA & 62.0 & 51.8 & 83.5 & 77.1 & .911 & .969 \\
MCCA (projection) & 60.1 & 50.6 & 82.7 & 76.4 & 1.0 & 1.0  \\
MCCA (concatenation) & 64.3 & 53.2 & 85.1 & 78.4 & .920 & .958  \\
\bottomrule
\end{tabular}
\end{sc}
\end{small}
\end{center}
\vskip -0.1in
\end{table}

\subsubsection{Friend recommendation}
\label{tab:friend_recommendation}

In a similar setting, we perform friend recommendation \cite{benton2016learning}. 500 user accounts were set aside (250 for each dev and test), and it was recorded which of the dev and test users followed those accounts. The friend recommendation task amounts to predicting which user accounts a given user is likely to follow. The train, dev and test set contain respectively 6522, 82608 and 81985 users. The task setup is identical to user engagement prediction. We sweep over the same values of hyperparameters as for user engagement prediction.

We report our results in Table~\ref{tab:twitter_friend}. Based on the results using the raw features, this task is much harder than user engagement prediction, described in \ref{subsec:user_engagement_prediction}. In this setting, MCCA performs better or on par with VCCA. The smaller performance gap between VCCA and MCCA could be explained by the much smaller size of the train set, and possibly the difficulty of obtaining a coherent clustering. For each of the best performing MCCA instances, the clustering at training time shows that one of the clusters collapses to 1, 2 or 3 points. For each method, the best performing hyperparameters vary from one task to the other, without any pattern emerging. Again, this shows the impact of task-based hyperparameter selection.

While the task is quite different from user engagement prediction described in Section~\ref{subsec:user_engagement_prediction}, the conclusions regarding VCCA and MCCA remain mostly identical.

\begin{table}[t]
\caption{Friend recommendation results on Twitter dataset. REC: Recall@1000, ROC-AUC: area under the ROC curve, MRR: mean reciprocal rank.}
\label{tab:twitter_friend}
\vskip 0.15in
\begin{center}
\begin{small}
\begin{sc}
\begin{tabular}{ p{2cm} p{0.5cm} p{0.5cm} p{0.5cm} p{0.5cm} p{0.5cm} p{0.5cm} }
\toprule
& \multicolumn{2}{c}{REC} & \multicolumn{2}{c}{ROC-AUC} & \multicolumn{2}{c}{MRR} \\ 
\cmidrule(r){2-3}\cmidrule(r){4-5}\cmidrule(r){6-7}
features & dev & test & dev & test & dev & test \\
\midrule
raw features & 3.21 & 2.96 & 59.6 & 59.3 & .821 & .788 \\
VCCA & 4.02 & 4.04 & 62.3 & 62.1 & .924 & .909 \\
MCCA (projection) & 3.83 & 3.48 & 62.2 & 61.8 & .994 & .996 \\
MCCA (concatenation) & 4.03 & 4.05 & 62.6 & 62.0 & .957 & .942 \\
\bottomrule
\end{tabular}
\end{sc}
\end{small}
\end{center}
\vskip -0.1in
\end{table}

\section{Conclusion}

In this paper, we proposed a novel objective for representation learning, combining CCA and mixture models, in conjunction with simple heuristics to maximize the objective at training time, and use the representations at test time. Evaluating our representations in different settings, we have shown the usefulness of both our objective and our heuristics. Overall, across tasks, our proposed method performs on par or better than the standard version of CCA.

Our results suggest that the proposed method is a valid objective, potentially a good generalization of the standard CCA objective. With more hyperparameters than the standard CCA objective, and the possibility of informing the mixture components with hierarchical structure present in the data, it has the potential to better adapt to downstream tasks. It would further benefit from a more thorough optimization scheme with provable guarantees, and possibly extensions to non-linear methods.

\bibliography{example_paper}

\begin{thebibliography}{31}
\providecommand{\natexlab}[1]{#1}
\providecommand{\url}[1]{\texttt{#1}}
\expandafter\ifx\csname urlstyle\endcsname\relax
  \providecommand{\doi}[1]{doi: #1}\else
  \providecommand{\doi}{doi: \begingroup \urlstyle{rm}\Url}\fi

\bibitem[Andrew et~al.(2013)Andrew, Arora, Bilmes, and Livescu]{andrew2013deep}
Andrew, Galen, Arora, Raman, Bilmes, Jeff, and Livescu, Karen.
\newblock Deep canonical correlation analysis.
\newblock In \emph{Int. Conf. on Machine Learning}, pp.\  1247--1255, 2013.

\bibitem[Arora \& Livescu(2013)Arora and Livescu]{arora2013multi}
Arora, Raman and Livescu, Karen.
\newblock Multi-view {CCA}-based acoustic features for phonetic recognition
  across speakers and domains.
\newblock In \emph{2013 IEEE Int. Conf. on Acoustics, Speech and Sig. Proc.},
  pp.\  7135--7139. IEEE, 2013.

\bibitem[Arora \& Livescu(2014)Arora and Livescu]{arora2014multi}
Arora, Raman and Livescu, Karen.
\newblock Multi-view learning with supervision for transformed bottleneck
  features.
\newblock In \emph{2014 IEEE Int. Conf. on Acoustics, Speech, and Sig. Proc.
  (ICASSP)}, pp.\  2499--2503. IEEE, 2014.

\bibitem[Bach \& Jordan(2005)Bach and Jordan]{bach2005probabilistic}
Bach, Francis~R and Jordan, Michael~I.
\newblock A probabilistic interpretation of canonical correlation analysis.
\newblock 2005.

\bibitem[Benton et~al.(2016)Benton, Arora, and Dredze]{benton2016learning}
Benton, Adrian, Arora, Raman, and Dredze, Mark.
\newblock Learning multiview embeddings of twitter users.
\newblock In \emph{Proc. 54th Annual Meeting of the Assoc. Comp. Linguistics
  (Volume 2: Short Papers)}, volume~2, pp.\  14--19, 2016.

\bibitem[Benton et~al.(2017)Benton, Khayrallah, Gujral, Reisinger, Zhang, and
  Arora]{benton2017deep}
Benton, Adrian, Khayrallah, Huda, Gujral, Biman, Reisinger, Dee~Ann, Zhang,
  Sheng, and Arora, Raman.
\newblock Deep generalized canonical correlation analysis.
\newblock \emph{arXiv preprint arXiv:1702.02519}, 2017.

\bibitem[Blei et~al.(2003)Blei, Ng, and Jordan]{blei2003latent}
Blei, David~M, Ng, Andrew~Y, and Jordan, Michael~I.
\newblock Latent dirichlet allocation.
\newblock \emph{Journal of machine Learning research}, 3\penalty0
  (Jan):\penalty0 993--1022, 2003.

\bibitem[Fawcett(2004)]{fawcett2004roc}
Fawcett, Tom.
\newblock {ROC} graphs: Notes and practical considerations for researchers.
\newblock \emph{Machine learning}, 31\penalty0 (1):\penalty0 1--38, 2004.

\bibitem[Fern et~al.(2005)Fern, Brodley, and Friedl]{fern2005correlation}
Fern, Xiaoli~Z, Brodley, Carla~E, and Friedl, Mark~A.
\newblock Correlation clustering for learning mixtures of canonical correlation
  models.
\newblock In \emph{Proceedings of the 2005 SIAM Int. Conf. on Data Mining},
  pp.\  439--448. SIAM, 2005.

\bibitem[Gales et~al.(2008)Gales, Young, et~al.]{gales2008application}
Gales, Mark, Young, Steve, et~al.
\newblock The application of hidden markov models in speech recognition.
\newblock \emph{Foundations and Trends{\textregistered} in Sig. Proc.},
  1\penalty0 (3):\penalty0 195--304, 2008.

\bibitem[Hardoon et~al.(2004)Hardoon, Szedmak, and
  Shawe-Taylor]{hardoon2004canonical}
Hardoon, David~R, Szedmak, Sandor, and Shawe-Taylor, John.
\newblock Canonical correlation analysis: An overview with application to
  learning methods.
\newblock \emph{Neural computation}, 16\penalty0 (12):\penalty0 2639--2664,
  2004.

\bibitem[Holzenberger et~al.(2019)Holzenberger, Palaskar, Madhyastha, Metze,
  and Arora]{holzenberger2019learning}
Holzenberger, Nils, Palaskar, Shruti, Madhyastha, Pranava, Metze, Florian, and
  Arora, Raman.
\newblock Learning from multiview correlations in open-domain videos.
\newblock In \emph{IEEE Int. Conf. Acoustics, Speech, and Sig. Proc. (ICASSP)},
  2019.

\bibitem[Horst(1961)]{horst1961generalized}
Horst, Paul.
\newblock Generalized canonical correlations and their applications to
  experimental data.
\newblock \emph{Journal of Clinical Psychology}, 17\penalty0 (4):\penalty0
  331--347, 1961.

\bibitem[Hosino(2010)]{hosino2010high}
Hosino, Tikara.
\newblock High dimensional non-linear modeling with bayesian mixture of {CCA}.
\newblock In \emph{Int. Conf. on Neural Information Processing}, pp.\
  446--453. Springer, 2010.

\bibitem[Hotelling(1936)]{hotelling1936relations}
Hotelling, Harold.
\newblock Relations between two sets of variates.
\newblock \emph{Biometrika}, 28\penalty0 (3/4):\penalty0 321--377, 1936.

\bibitem[Klami \& Kaski(2007)Klami and Kaski]{klami2007local}
Klami, Arto and Kaski, Samuel.
\newblock Local dependent components.
\newblock In \emph{Proceedings of the 24th Int. Conf. on Machine learning},
  pp.\  425--432. ACM, 2007.

\bibitem[Koehn \& Knowles(2017)Koehn and Knowles]{koehn2017six}
Koehn, Philipp and Knowles, Rebecca.
\newblock Six challenges for neural machine translation.
\newblock \emph{arXiv preprint arXiv:1706.03872}, 2017.

\bibitem[Kuhn et~al.(2000)Kuhn, Junqua, Nguyen, and Niedzielski]{kuhn2000rapid}
Kuhn, Roland, Junqua, J-C, Nguyen, Patrick, and Niedzielski, Nancy.
\newblock Rapid speaker adaptation in eigenvoice space.
\newblock \emph{IEEE Trans. Speech and Audio Proc.}, 8\penalty0 (6), 2000.

\bibitem[Lai \& Fyfe(2000)Lai and Fyfe]{lai2000kernel}
Lai, Pei~Ling and Fyfe, Colin.
\newblock Kernel and nonlinear canonical correlation analysis.
\newblock \emph{International Journal of Neural Systems}, 10\penalty0
  (05):\penalty0 365--377, 2000.

\bibitem[Malkov \& Yashunin(2018)Malkov and Yashunin]{malkov2018efficient}
Malkov, Yury~A and Yashunin, Dmitry~A.
\newblock Efficient and robust approximate nearest neighbor search using
  hierarchical navigable small world graphs.
\newblock \emph{IEEE Transactions on Pattern Analysis and Machine
  Intelligence}, 2018.

\bibitem[Ngiam et~al.(2011)Ngiam, Khosla, Kim, Nam, Lee, and
  Ng]{ngiam2011multimodal}
Ngiam, Jiquan, Khosla, Aditya, Kim, Mingyu, Nam, Juhan, Lee, Honglak, and Ng,
  Andrew~Y.
\newblock Multimodal deep learning.
\newblock In \emph{Proc. Int. Conf. Machine Learning}, 2011.

\bibitem[Podosinnikova et~al.(2016)Podosinnikova, Bach, and
  Lacoste-Julien]{podosinnikova2016beyond}
Podosinnikova, Anastasia, Bach, Francis, and Lacoste-Julien, Simon.
\newblock Beyond {CCA}: moment matching for multi-view models.
\newblock \emph{arXiv:1602.09013}, 2016.

\bibitem[Rastogi et~al.(2015)Rastogi, Van~Durme, and
  Arora]{rastogi2015multiview}
Rastogi, Pushpendre, Van~Durme, Benjamin, and Arora, Raman.
\newblock Multiview {LSA}: Representation learning via generalized {CCA}.
\newblock In \emph{Proceedings of the 2015 Conference of the North American
  Chapter of the Association for Computational Linguistics: Human Language
  Technologies}, pp.\  556--566, 2015.

\bibitem[Sugamura et~al.(1983)Sugamura, Shikano, and
  Furui]{sugamura1983isolated}
Sugamura, Noboru, Shikano, Kiyohiro, and Furui, Sadaoki.
\newblock Isolated word recognition using phoneme-like templates.
\newblock In \emph{Acoustics, Speech, and Sig. Proc., IEEE Int. Conf. on
  ICASSP'83.}, volume~8, pp.\  723--726. IEEE, 1983.

\bibitem[V{\'a}squez-Correa et~al.(2017)V{\'a}squez-Correa, Orozco-Arroyave,
  Arora, N{\"o}th, Dehak, Christensen, Rudzicz, Bocklet, Cernak, Chinaei,
  et~al.]{vasquez2017multi}
V{\'a}squez-Correa, Juan~Camilo, Orozco-Arroyave, Juan~Rafael, Arora, Raman,
  N{\"o}th, Elmar, Dehak, Najim, Christensen, Heidi, Rudzicz, Frank, Bocklet,
  Tobias, Cernak, Milos, Chinaei, Hamidreza, et~al.
\newblock Multi-view representation learning via {GCCA} for multimodal analysis
  of parkinson's disease.
\newblock In \emph{2017 IEEE Int. Conf. on Acoustics, Speech, and Sig. Proc.
  (ICASSP)}, pp.\  2966--2970. IEEE, 2017.

\bibitem[Viinikanoja et~al.(2010)Viinikanoja, Klami, and
  Kaski]{viinikanoja2010variational}
Viinikanoja, Jaakko, Klami, Arto, and Kaski, Samuel.
\newblock Variational bayesian mixture of robust {CCA} models.
\newblock In \emph{Joint European Conference on Machine Learning and Knowledge
  Discovery in Databases}, pp.\  370--385. Springer, 2010.

\bibitem[Wang(2007)]{wang2007variational}
Wang, Chong.
\newblock Variational bayesian approach to canonical correlation analysis.
\newblock \emph{IEEE Transactions on Neural Networks}, 18\penalty0
  (3):\penalty0 905--910, 2007.

\bibitem[Wang et~al.(2014)Wang, Arora, and Livescu]{wang2014reconstruction}
Wang, Weiran, Arora, Raman, and Livescu, Karen.
\newblock Reconstruction of articulatory measurements with smoothed low-rank
  matrix completion.
\newblock In \emph{Spoken Language Technology Workshop (SLT), 2014 IEEE}, pp.\
  54--59. IEEE, 2014.

\bibitem[Wang et~al.(2015{\natexlab{a}})Wang, Arora, Livescu, and
  Bilmes]{wang2015deep}
Wang, Weiran, Arora, Raman, Livescu, Karen, and Bilmes, Jeff.
\newblock On deep multi-view representation learning.
\newblock In \emph{Int. Conf. on Machine Learning}, pp.\  1083--1092,
  2015{\natexlab{a}}.

\bibitem[Wang et~al.(2015{\natexlab{b}})Wang, Arora, Livescu, and
  Bilmes]{wang2015unsupervised}
Wang, Weiran, Arora, Raman, Livescu, Karen, and Bilmes, Jeff~A.
\newblock Unsupervised learning of acoustic features via deep canonical
  correlation analysis.
\newblock In \emph{Acoustics, Speech, and Sig. Proc. (ICASSP), 2015 IEEE Int.
  Conf. on}, pp.\  4590--4594. IEEE, 2015{\natexlab{b}}.

\bibitem[Westbury et~al.(1990)Westbury, Milenkovic, Weismer, and
  Kent]{westbury1990x}
Westbury, John, Milenkovic, Paul, Weismer, Gary, and Kent, Raymond.
\newblock X-ray microbeam speech production database.
\newblock \emph{The Journal of the Acoustical Society of America}, 88\penalty0
  (S1):\penalty0 S56--S56, 1990.

\end{thebibliography}
\bibliographystyle{icml2018}

\end{document}